\documentclass[sigconf]{acmart}

\usepackage{booktabs}
\usepackage{enumerate}
\usepackage{enumitem}
\usepackage{balance}

\copyrightyear{2019}
\acmYear{2019} 
\setcopyright{iw3c2w3}
\acmConference[WWW '19]{Proceedings of the 2019 World Wide Web Conference}{May 13--17, 2019}{San Francisco, CA, USA}
\acmBooktitle{Proceedings of the 2019 World Wide Web Conference (WWW '19), May 13--17, 2019, San Francisco, CA, USA}
\acmPrice{}
\acmDOI{10.1145/3308558.3313399}
\acmISBN{978-1-4503-6674-8/19/05}

\begin{document}
\title{Generating Titles for Web Tables}

\author{Braden Hancock}\thanks{*BH completed the majority of this work while at Google.}
\affiliation{%
  \institution{*Stanford University}
  \streetaddress{353 Serra Mall}
  \city{Stanford} 
  \state{CA} 
  \postcode{94305}
}
\email{bradenjh@cs.stanford.edu}

\author{Hongrae Lee}
\affiliation{%
  \institution{Google}
  \streetaddress{1950 Charleston Rd}
  \city{Mountain View} 
  \state{CA}
}
\email{hrlee@google.com}

\author{Cong Yu}
\affiliation{%
  \institution{Google}
  \streetaddress{111 8th Ave}
  \city{New York} 
  \state{NY} 
  \postcode{10011}
}
\email{congyu@google.com}

% If the default list of authors is too long for headers.
% \renewcommand{\shortauthors}{B. Hancock et al.}

\begin{abstract}
Descriptive titles provide crucial context for interpreting tables that are extracted from web pages and are a key component of search features such as tabular featured snippets from Google and Bing. Prior approaches have attempted to produce titles by selecting existing text snippets associated with the table. These approaches, however, are limited by their dependence on suitable titles existing \textit{a priori}. In our user study, we observe that the relevant information for the title tends to be scattered across the page, and often---more than 80\% of the time---does not appear verbatim anywhere in the page. We propose instead the application of a sequence-to-sequence neural network model as a more generalizable approach for generating high-quality table titles. This is accomplished by extracting many text snippets that have potentially relevant information to the table, encoding them into an input sequence, and using both copy and generation mechanisms in the decoder to balance relevance and readability of the generated title. We validate this approach with human evaluation on sample web tables and report that while sequence models with only a copy mechanism or only a generation mechanism are easily outperformed by simple selection-based baselines, the model with both capabilities performs the best, approaching the quality of crowdsourced titles while training on fewer than ten thousand examples. To the best of our knowledge, the proposed technique is the first to consider text-generation methods for table titles, and establishes a new state of the art.
\end{abstract}


\maketitle

\section{Introduction}
\label{sec:intro}

\begin{figure}[t]
  \centering
    \begin{minipage}[c]{0.5\columnwidth}
      \centering
      \includegraphics[width=0.95\columnwidth]{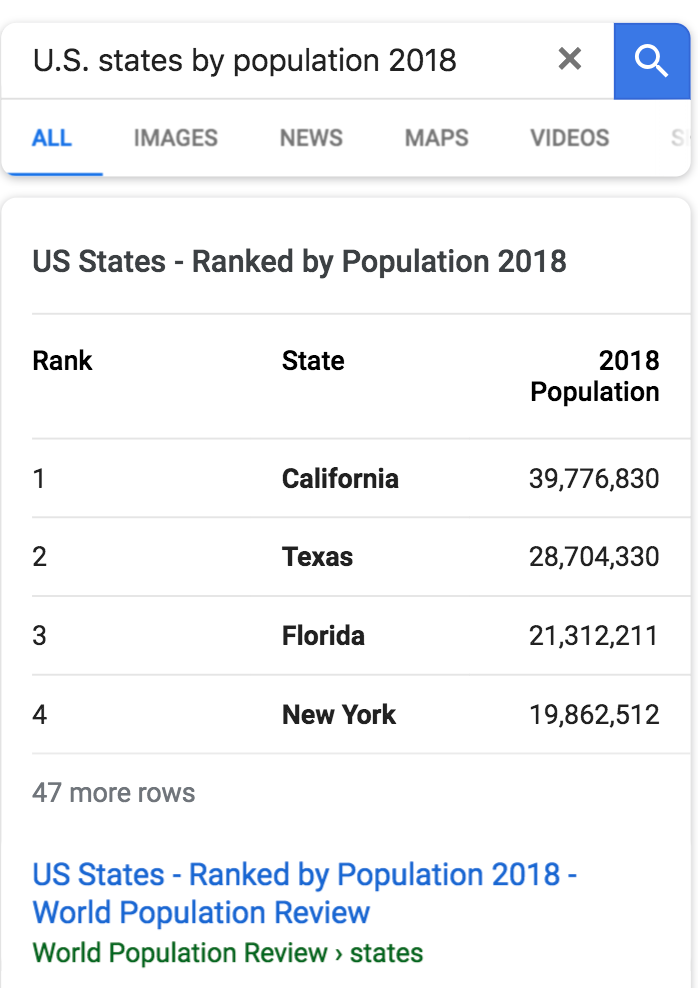}
      \label{fig:test1}
    \end{minipage}%
    \begin{minipage}[c]{0.5\columnwidth}
      \centering
      \includegraphics[width=0.9\columnwidth]{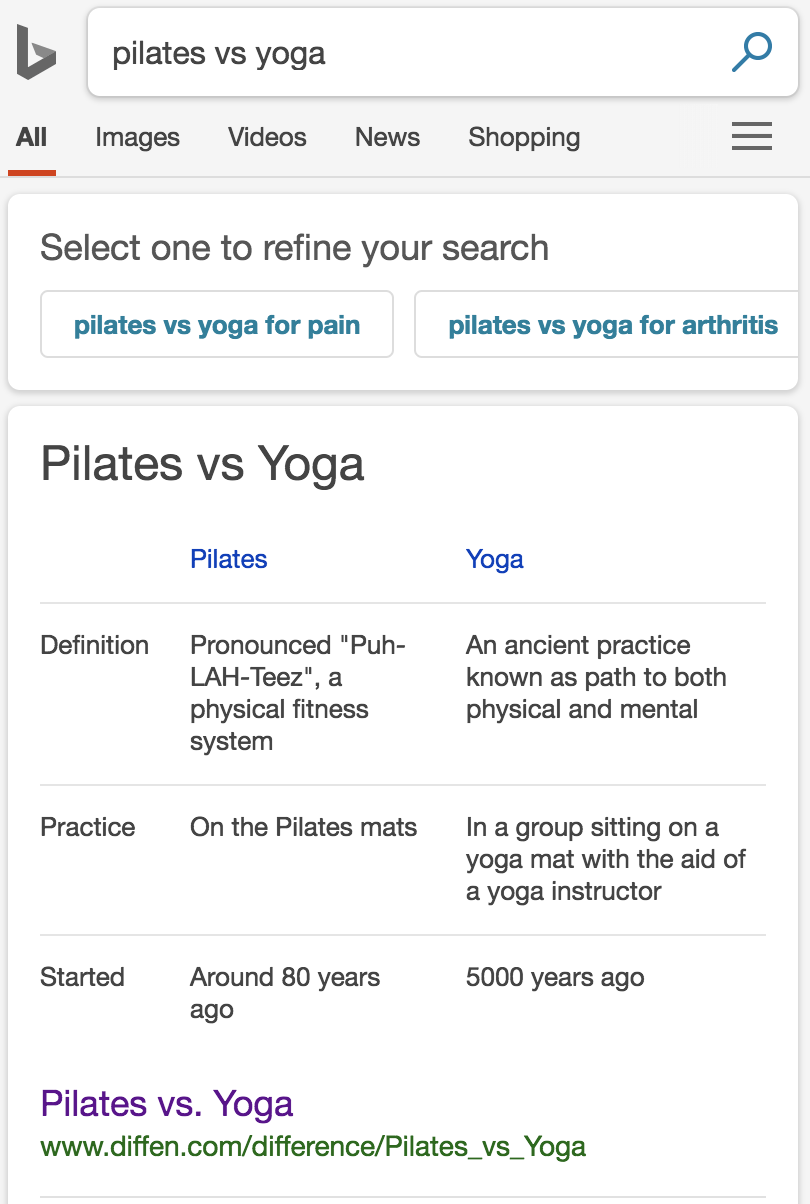}
      \label{fig:test2}
    \end{minipage}  
  \caption{Example tabular results in search engines}
  \label{fig:table_answers}
\end{figure}

Modern search engines no longer simply return links to relevant web pages. Where possible, they enhance the search experience by extracting and directly displaying the information that a user is seeking. Often, this information is best displayed in a semi-structured form, such as a table or list; for example, in response to the query ``U.S. states by population 2018,'' a relevant table showing population by state is more effective than a sentence with the same information. It is now a major feature in mainstream search engines, e.g., ~\cite{bing2017answers, sullivan2017answers} and Figure~\ref{fig:table_answers} shows examples of tabular results in Google and Bing.

However, separating a table from its original context removes important clues that help a user interpret its contents and trust its relevance. In our experience, the most effective way of providing this crucial context is by accompanying the tables with corresponding descriptive titles.

\begin{figure}[t]
  \centering
  \vspace{0.25in}
  \includegraphics[width=3in]{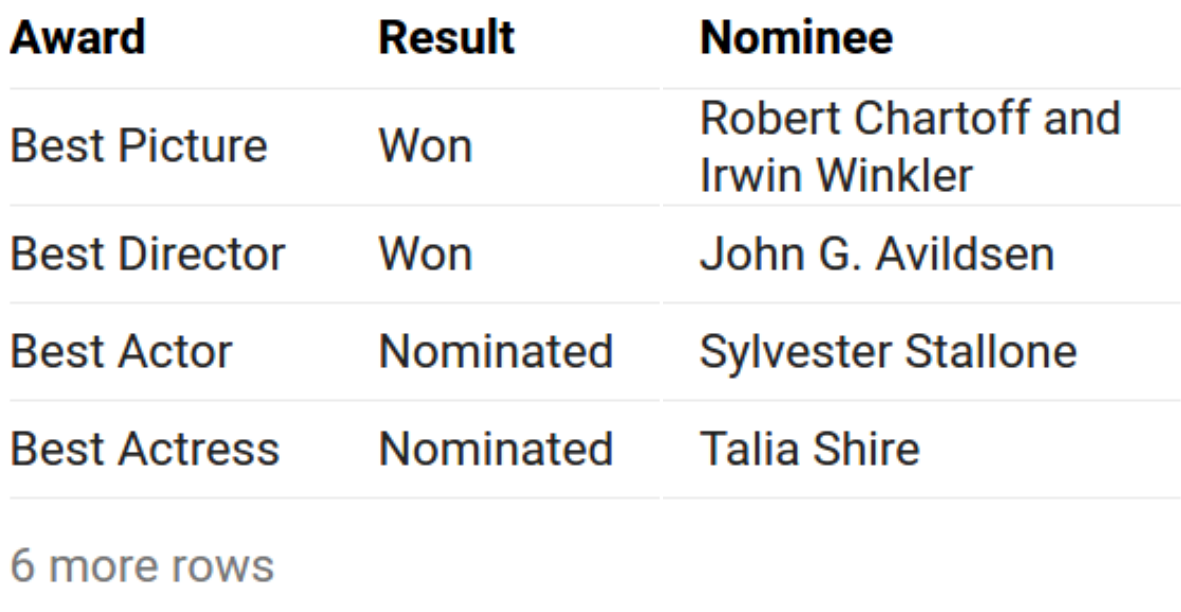}
  \caption{A descriptive title provides crucial context for understanding the contents of a table when it is displayed out of its original context. Consider, for example, the experience of trying to interpret or assess the relevance of this extracted table with versus without its title: ``\emph{Rocky} Academy Award Nominations (1977).''}
  \label{fig:rocky}
\end{figure}

Consider, for example, the table shown in Figure~\ref{fig:rocky}. If this table were shown in isolation, it would be difficult to say for certain what it is about. If the search query that led to this table being returned is known (e.g., ``Rocky Franchise Awards''), this helps with interpretation but still leaves many questions unanswered, such as: Does this table provide the exact information I was looking for or is it just the most relevant table that could be found? Are the awards in the table for one \emph{Rocky} film or the entire \emph{Rocky} franchise? Are these all from the same award source or from many sources? When a descriptive title is added, understanding of the table's contents improves, as does trust in the table's relevance to the query. The goal is for the user to have enough information from the displayed result (including the title) to know whether or not it is relevant to his or her information need.

Beyond user experience, the ability to generate titles for or otherwise summarize the content of semi-structured data enables other use cases as well. In a search engine, it plays a crucial role in supporting web applications such as table snippets~\cite{balakrishnan@cidr15}.
From an information extraction standpoint, titles generated offline may be used as an additional signal when assessing the relevance of the summarized content with respect to user queries. With the increasing popularity of hands-free interfaces such as Google Home \citep{googlehome} and Amazon Echo \citep{amazonecho}, descriptive titles also provide a way for such devices to summarize a table's content in a more understandable way than reading the data row-by-row.

\begin{figure*}[t]
  \centering
  \includegraphics[width=7in]{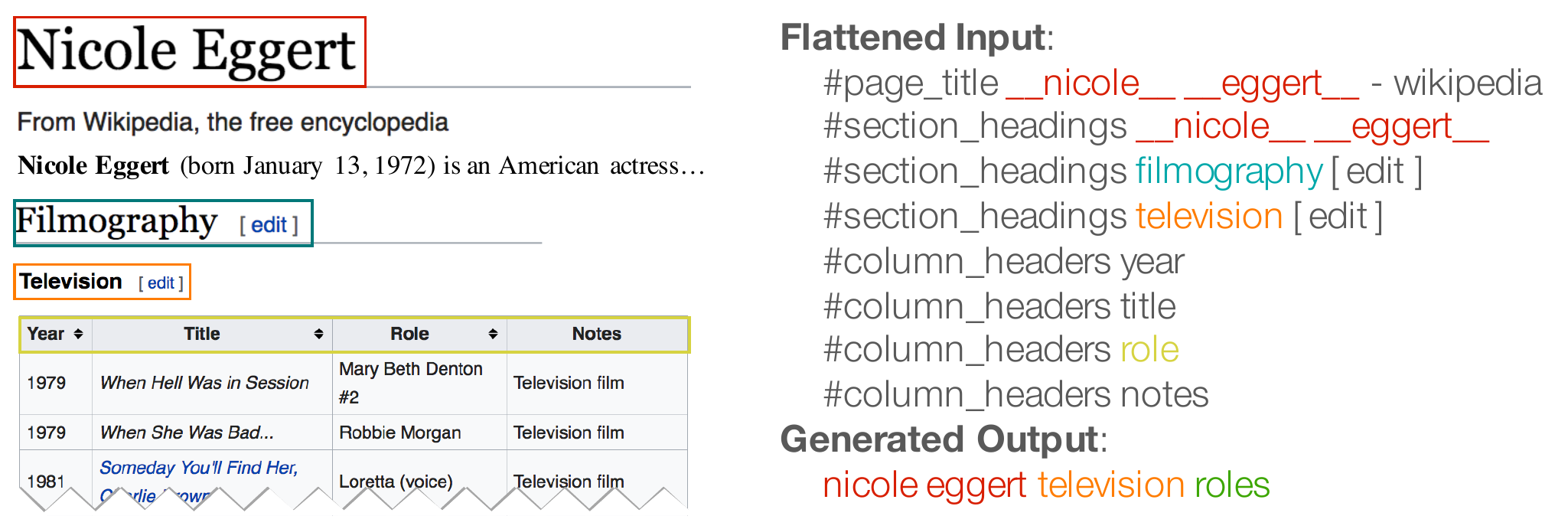}
  \caption{On the left is the top portion of a table from the web along with preceding section headings (some content has been removed for clarity). On the right are the input string (split into multiple lines for readability) and output of the trained title generation network. The double underscores around ``nicole'' and ``eggert'' denote that these are out-of-vocab (OOV) tokens (i.e., they were not present in the training data, and can therefore only be copied, not generated). Note that the generated title is composed from multiple fields (but not all fields), irrelevant tokens such as ``[edit]'' and ``- wikipedia'' are not copied, and while ``role'' may help to suggest the schema of this table, the token ``roles'' was actually generated, reflecting a learned pattern that titles tend to refer to plural objects.}
  \label{fig:nicole}
\end{figure*}

\subsection{Challenges}

In practice, automatically producing descriptive titles is quite challenging. First of all, the semantics of a table tend to be distributed among a variety of elements on a web page. For example, consider the table shown in Figure~\ref{fig:nicole}: the page title includes the name of the person being discussed, the section headings describe the topic of the table, and the column headers give the schema of the table contents. Any single one of these text strings would be insufficiently precise to serve as a table title, but together they include all the necessary information for a human to interpret its contents. This raises the question of how to best represent this collection of text snippets for machine learning.

Second, a title must sound natural; i.e., it should sound like something a human would write and not a ``keyword soup.'' Clearly for the example in Figure~\ref{fig:nicole}, simply concatenating the text from each of the table metadata components would result in a title with poor readability.

Third, tables tend to describe specific relationships between specific entities, much more so than unstructured data such as free text. Consequently, descriptive titles for tables tend to include many proper nouns and rare or out-of-vocab (OOV) tokens, which are well-documented stumbling blocks for natural language generation algorithms \cite{bahdanau2014neural, luong2014addressing, sutskever2014sequence}.

Finally, the variety of tables on the web is immense \citep{crestan2011web,lehmberg2016large}. Some tables are relational (``horizontal'') while others are entity-based (``vertical'')~\cite{balakrishnan@cidr15}. A table may have five rows or 50. It may have an explicit caption and three section headings or none of either. It may contain the primary content of a web page or be purely supplemental, or it may be used entirely for formatting purposes and not actually be expressing any relationship at all. This tremendous diversity renders many heuristic-based approaches inadequate, as evidenced in part by the unsuccessful attempts to heuristically bootstrap training data which we report in Section~\ref{sec:datasets}.

Ultimately, a generated title must satisfy two properties: relevance and readability. A \textbf{relevant} title is one which correctly describes what information the table contains; it should be neither too vague (``Scores'') nor too specific (``The box score of the baseball game that the Houston Astros won against the Los Angeles Dodgers in 10 innings in Game 5 of the 2017 World Series''). A \textbf{readable} title is one which sounds natural to a human reader. It need not be a complete sentence (indeed, most natural titles are not), but it should be fairly straightforward for a human to parse.

\subsection{Conventional Approaches}

To address readability, a common approach is to treat the title production task as a selection problem--given all the text snippets on the page, identify the one most likely to make a good title for a table in question. By selecting existing text, such approaches avoid the complexity of learning a language model for concatenating words and phrases in a natural way. However, to succeed, these approaches require a high quality title to appear somewhere verbatim on the web page, which is rarely the case. As will be discussed in Section~\ref{sec:datasets}, in more than 80\% of cases in our user study, users did not consider any extracted text snippets as the best representation of the table. The caption tag, which is supposed to provide a title of the table, is used in widely different ways across table authors, and moreover, is present in only less than a few percentage of web tables.

When existing text snippets \emph{are} relevant to a table, they often each contain only a portion of the ideal title for that table. Consider, for example, the titles shown in Figure~\ref{fig:nicole} and Figure~\ref{fig:compositionality}---all three titles contain text from at least three separate locations on the page. This highlights the need for compositionality in any title generation approach, which most IR-based approaches do not support.

Another approach for title production would be to mine query logs for past user queries that led to that table and use these as candidate titles \cite{chirigati2016knowledge}. Using human-generated text increases the likelihood of the title being compositional (since it is not restricted to text snippets from the page), but once again, the success of the approach is dependent on a high quality title existing \textit{a priori} for selection. For tables with few corresponding queries, success is unlikely. Furthermore, for web pages with multiple tables in them, the task of linking a query that led to that page with the most relevant table on the page (if any) is non-trivial.

\subsection{Proposed Approach}

In this paper, we propose a new framework for generating high quality table titles. Our approach utilizes a sequence-to-sequence neural network model with both a copy mechanism and a generation mechanism. By generating titles instead of selecting or ranking existing text strings, this approach is capable of composing high quality title strings even when none exist in the source text. At the same time, rare and OOV tokens are retrievable thanks to the copy mechanism (see Figure~\ref{fig:nicole}, for example). Furthermore, as a high-capacity machine learning model, with sufficient training data the model is able to learn and utilize more nuanced patterns than most heuristic-based approaches, making it more robust to the variety of tables. 

The contributions of this paper are as follows:
\begin{enumerate}[topsep=2pt]
 \item We identify the key components for successful table title generation (semi-structured data representation, composition from multiple sources, and handling rare and OOV tokens) and propose techniques for dealing with each in the context of web tables.
 \item We explore qualitatively and quantitatively the value of having both copy and generation mechanisms in a model for table title generation, finding that both are necessary to surpass simple baseline performance.
 \item We propose a framework for generating high quality titles for web tables and find that both by automated metrics (ROUGE scores) and human evaluation on a held-out test set, our model outperforms selection-based baselines and sequence-to-sequence models with either a generation mechanism or a copy mechanism, falling just slightly below human-generated titles on relevance and readability.
\end{enumerate}

\section{Web Table Title Generation}
\label{sec:approach}

\begin{table*}
    \centering
    \begin{tabular}{ll}
      \toprule
      Field     & Description \\
      \midrule
      Page title* & Tokens inside the <title> tag nested in the <head> tag of the web page \\
      Section headings* & Tokens in <h1>, <h2>, etc. tags with increasing priority, starting with the nearest \\
      Table captions* & Captions inside <caption> tags \\
      Spanning headers* & Headers in <th> tags that span all columns \\
      Column headers* & Headers in <th> tags \\
      Prefix text & Up to 200 tokens preceding the table until a new table or section boundary\\
      Suffix text & Up to 200 tokens following the table until a new table or section boundary\\
      Table rows & Text inside a <tr> tag, comma delimited with other tags (e.g., <td>) removed\\
      \bottomrule
    \end{tabular}
    \vspace{1ex}
    \bigskip
    \caption{Metadata fields considered for inclusion in the input to the title generation model. Field names followed by * were included for the experiments reported in Section~\ref{sec:experiments}. Where multiple values are found, we maintain the list of values in reading order.}
    \label{tab:fields}
  \end{table*}

\subsection{Web Tables}

The web is filled with tables. In a seminal 2008 study, researchers at Google extracted 14.1 billion HTML tables from Google's general-purpose web crawl and released initial findings on the properties and challenges associated with this corpus \cite{cafarella2008webtables}. Since that publication, hundreds of papers have been published on various applications associated with web tables, including table detection, table classification, relation extraction, knowledge base construction, data cleaning, query answering using tables, and many others~\cite{limaye2010vldb,YakoutGCC12,pimplikar@vldb2012,Yu12,wang@www15,YangDCC14,balakrishnan@cidr15}. To this important body of work we add our own contribution of title generation for web tables. 

Figure~\ref{fig:nicole} shows excerpts from the top portion of a web page that has a table in it. As a human, it is easy to recognize that this table shows television appearances by Nicole Eggert. Note, however, that only one of the words in the target title for this table actually occurs within the table. Instead, the meaning of the table is distributed among many different elements on the page. This raises the important question of what context should be considered when creating a title for a table. 

One option would be to include the entire web page. This would be guaranteed to include any text that could possibly be relevant for titling a table; unfortunately, it would almost certainly also include a significant amount of unrelated text, decreasing the signal-to-noise ratio of the input. The task of identifying pertinent information in a document is its own open research problem of significant complexity. Additionally, the sequential nature of encoding information with a sequence-to-sequence models means that training time is proportional to the length of the sequence being encoded. Computation overheads for encoding tens or hundreds of thousands of tokens as input is likely prohibitive for our task.

Empirically, we know that the most relevant information for titling a table tends to come from a relatively small set of potential text sources. One obvious source of such text is the contents of the \emph{caption} tag within a table element. However, less than $5\%$ of tables for which we would like to generate titles have explicit captions (including the one in Figure~\ref{fig:nicole}). Other relevant sources outside the table itself include the page title or section headings. While these certainly have greater coverage than captions, they also introduce new sources of ambiguity. Both page titles and section headings may be only loosely related to the content of an associated table or they may refer to multiple tables on the page which actually have important distinctions between them.

\begin{figure*}
  \centering
  \includegraphics[width=7in]{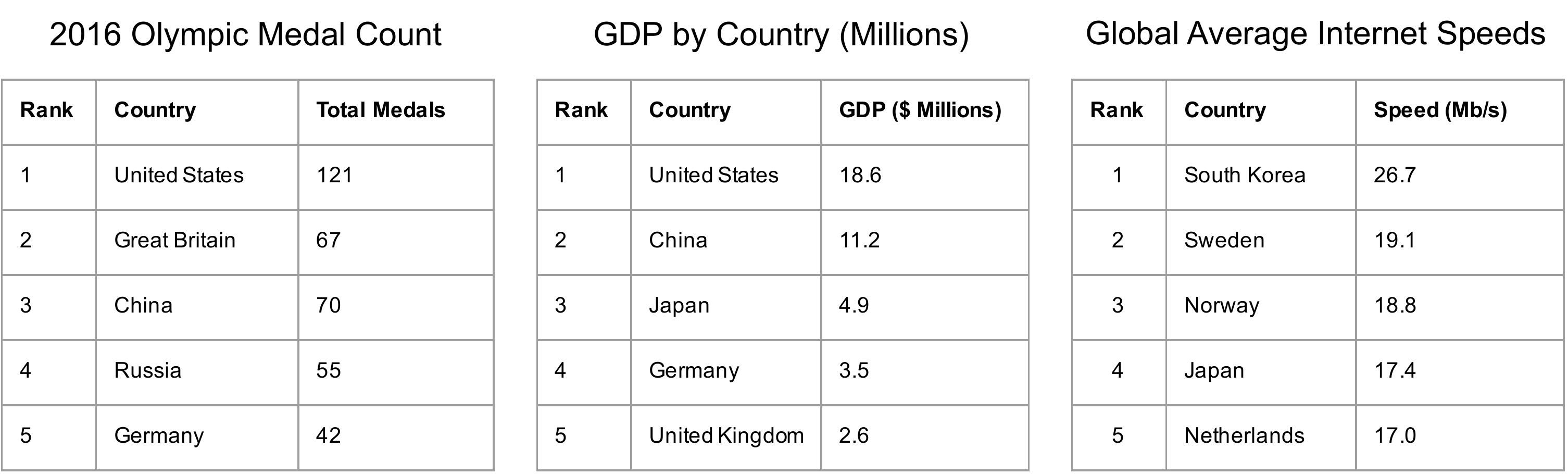}
  \caption{The contents of non-header rows of tables contribute surprisingly little information toward their titles tables. Consider these three tables with very similar non-header contents but very different topics and titles.}
  \label{fig:useless_contents}
\end{figure*}
In addition to these text snippets which tend to show up in their own set of HTML tags (e.g., <title> or <h1>), spans of unstructured text in sentences surrounding a table may also include important contextual clues. For example, in Figure~\ref{fig:nicole}, the first sentence on the page mentions that Nicole Eggert is an actress, which clearly pertains to the displayed table that lists television shows in which she has performed. While relevant text that alludes to the table may theoretically appear anywhere in the document, to reduce complexity and runtime, we use the heuristic of looking only at the unstructured text immediately preceding or following a table.

\subsection{Representing Semi-Structured Data}

\begin{figure*}[t]
  \centering
  \includegraphics[width=7in]{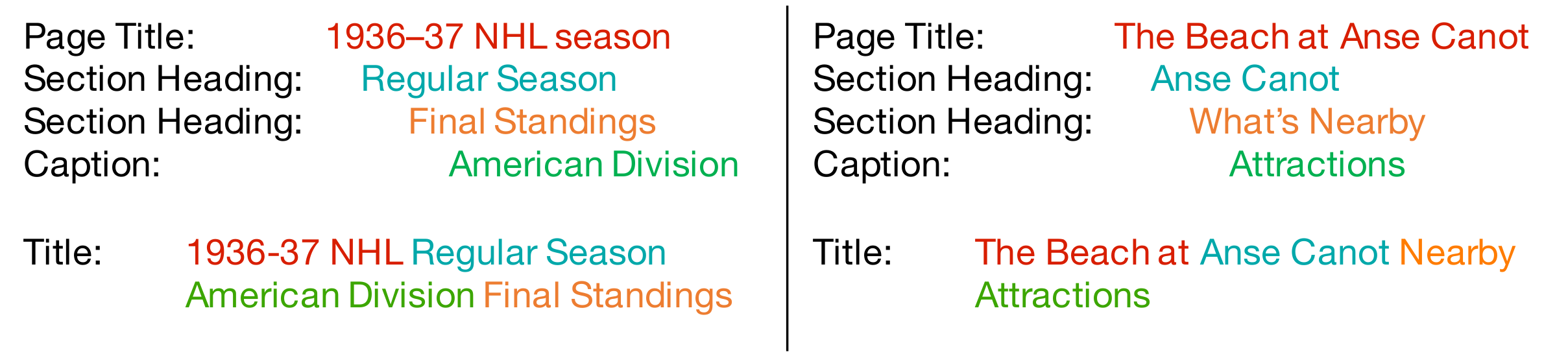}
  \caption{For the majority of tables, the ideal title is composed from multiple associated text snippets, rather than selected from among them.}
  \label{fig:compositionality}
\end{figure*}

Once we know what information we would like to include in the input to our model, the question arises of how to represent this information. Drawing on the rich and rapidly expanding body of work on encoder-decoder frameworks, we chose to model these structured fields as a sequence and found that table title generation lends itself well to such an approach. The output---a title in natural language---is clearly sequential in nature. And the input, a combination of textual elements associated with the table, exhibits a sequential nature on a couple different levels. First, the text within each field is ordered as it reads left to right. And second, as demonstrated in Figures~\ref{fig:nicole} and \ref{fig:compositionality}, the metadata fields that we collect for each table tend to increase in specificity as they get nearer to the table. Just as they were intended to be read in that order on the web page, they tend to remain in that order in high quality titles.

For each table in our dataset, we collected the following metadata fields (which are also listed in Table~\ref{tab:fields} and described in greater detail below): page title, section headings, table captions, spanning headers, column headers, prefix text, suffix text, and table rows (the contents of the table, one row at a time). Then, to convert these fields (key-value pairs) into a sequence, we iterated through the fields in the order just given and concatenated their contents, preceding each value (one or more tokens) with a special field token of the form ``\#[field name]''. The model treats these field tokens the same as any other, eventually learning to encode information differently based on what field it is reading and to never generate these field tokens in the decoder, since they never appear in the training data titles. We now describe in greater detail how each of these fields were collected.

\textbf{Page title}: Page titles are found inside the <title> tag in the <head> section of a web page. This title is not directly displayed on the web page in a browser, so many web pages repeat the page title as a section heading for the user to see (as in Figure~\ref{fig:nicole}). Given their applicability to the contents of the entire page, page titles proved to be one of the most consistently relevant fields for generated titles. While each page has only one page title, many page titles effectively concatenate multiple titles with delimiters, as in ``Chocolate Gifts | Artisan Truffles | Gourmet Chocolate''.

\textbf{Section headings}: We chose to rely primarily on explicit header tags (<h1>, <h2>, etc.) for section headings, using the numeric levels of the headers as proxies for parent-child and sibling relationships between sections. Rather than storing all headers preceding a table, we worked backward through the document, starting at the beginning of the table and ending at the beginning of the document. In this traversal, we saved only the first header (closest to the table) at each level that had higher priority (lower numeric value) than any headers collected so far (i.e., if a heading of type <h3> had already been observed, then all section headings of type <h\#> where \# $\geq 3$ were skipped from then on). Because not all web pages use these tags to denote headers, however, (relying instead on font size, bold formatting, centering, etc.), not all section headings that a human might identify were always collected. The problem of identifying document structure from visual information is an open research problem.

\textbf{Table captions}: While table captions (in explicit <caption> tags) tended to be highly relevant, they were relatively rare ($<5\%$\footnote{The statistics in the paper are from samples in our internal corpus unless specified.} of tables) and rarely sufficient as a title. Instead, captions often either provided the final piece of information in a long chain of text snippets that together form an adequate title (as in Figure~\ref{fig:compositionality}), or they represented a long caption (sometimes multiple sentences) as in, for example, scientific literature. Regardless, if a table had a caption, some portion of it tended to show up in both crowd-sourced and generated titles.

\textbf{Column headers}: Column headers are table cells denoted with <th> tags (instead of the <td> tags for table data). Most tables had only one row of column headers, but not all. Column headers tended to be short, often just one or two words each.

\textbf{Spanning headers}: Spanning headers are a special type of column header---one which spans all columns in the table. Some tables use these effectively as table titles, just placed inside the table instead of above it. Spanning headers were even rarer than captions, showing up in $<1\%$ of tables.

\textbf{Prefix/suffix text}: Prefix and suffix text are the tokens immediately before and after the table, respectively. They consisted of up to 200 tokens, stopping short at the boundary of another table or another section, since such boundaries tend to signal a change in topic. In practice, we found the prefix and suffix text to be quite noisy, rarely providing relevant title components not already included in the other fields, while significantly increasing the size of the input sequence. Consequently, these fields were not included in the dataset used to train the models for which we report results.

\textbf{Table rows}:
Table rows refer to contents of all non-header rows in a table. We were surprised at first to observe that including table rows actually hurt model performance, but upon closer inspection, we recognized that individual table records rarely include unique information that appears in a good title, and topical information expressed in the columns is generally captured by the column headers. Thus, the final dataset did not include the table rows field. Figure~\ref{fig:useless_contents} illustrates this finding with three tables that have very similar contents in their non-header rows, none of which contribute much to their respective titles.

\subsection{Supporting Compositionality}
\label{sec:compositionality}

Because of the inherent complexity of generating new text, our initial attempts to solve the problem of titling tables were selection-based. However, we soon found that the upper bound of such an approach was far too low, due to the frequency with which no adequate title could be found in a single continuous span of text. This observation was further confirmed during the collection of the crowdsourced dataset we describe in Section~\ref{sec:datasets}. For each human-generated title that we collected, we had the worker denote whether the optimal title that they generated occurred anywhere on the page in its entirety; 83\% of the time, it did not. 

As an illustration of this trend, consider the two collections of text snippets shown in Figure~\ref{fig:compositionality}. Each corresponds to a single table whose human-generated title is also given. For both tables, all four snippets are relevant, but it is only when these pieces are composed that a sufficiently correct title is created. 

\begin{figure*}
  \centering
  \includegraphics[width=7in]{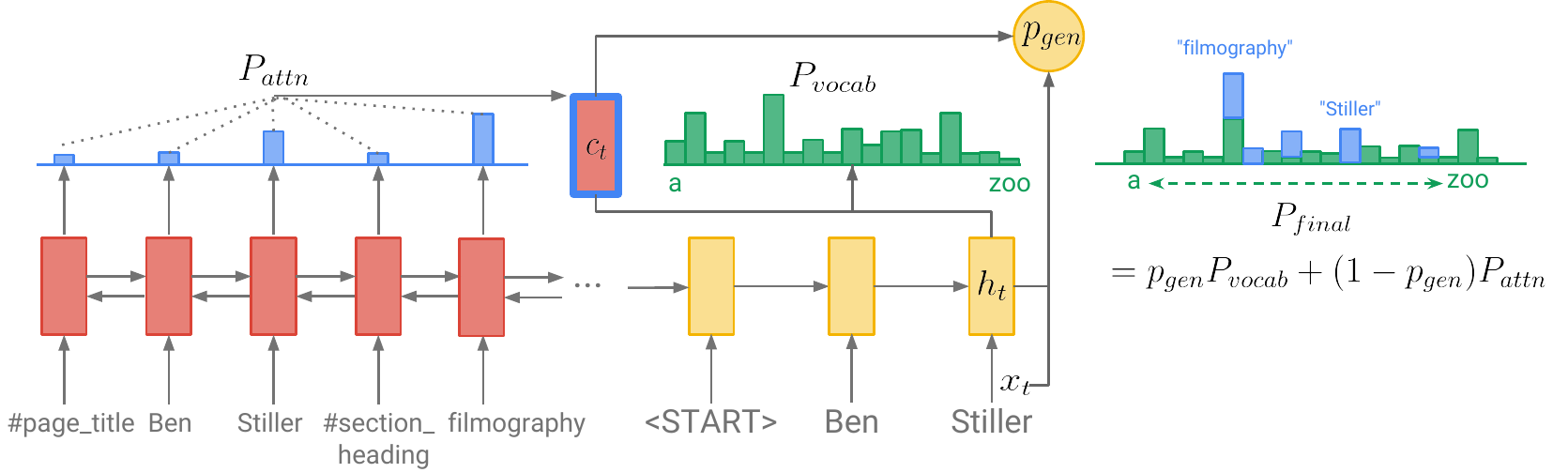}
  \caption{At each decode step, the pointer-generator network calculates the scalar value $p_{gen}$, which determines the relative contributions of the \emph{vocabulary distribution} (tokens that can be generated) and the \emph{attention distribution} (tokens that can be copied from the input sequence). Beam search is performed over the \emph{final distribution}.}
  \label{fig:pointer-generator}
\end{figure*}

\subsection{Handling Rare and OOV Tokens}
Successfully handling rare and OOV tokens is a challenge for most text generation tasks. This problem is exacerbated for the task of table title generation, due to the relational nature of tables; Tables tend to describe specific entities and quantities rather than abstract ideas. A generic description like "people shopping at an outdoor market" may work well for images, but generic titles like "countries and numbers" for tables in Figure~\ref{fig:useless_contents} would not be helpful. We observe the proportion of specific entities to generic terms is much higher in the very information-dense format of tables. For example, in the validation split of our crowdsourced dataset (approx. 1000 titles), 90\% of titles contained one or more proper nouns, and 45\% contained one or more OOV words (for a vocab based on over 8000 tables plus titles).

To overcome this stumbling block, we included a copy mechanism in the decoder of our sequence-to-sequence model. A copy mechanism allows a decoder network to not only generate tokens from a known vocabulary, but also occasionally copy a token directly from the input. The parameters controlling when a token is copied versus generated are also learned. Consequently, the model is able to learn specific patterns from the training data that suggest that the following word in a title is likely to be a rare or OOV word that should be copied rather than generated. For example, after producing the tokens ``Action Films Starring,'' with high probability, the next generated token will be the name of an actor or actress, of which there are many, and whose embeddings are likely very close to one another. However, that actor's or actress's name will almost certainly appear in the metadata associated with the table (e.g., in the page title or a section heading). In this case, copying over one of the unknown tokens that is being used like a name in the input (where ``being used like a name'' is a soft pattern the network naturally learns) is far more likely to succeed than generating a token.

\subsection{Model Selection}
The model architecture we used was a pointer-generator network, first introduced by \cite{see2017get} as a tool for abstractive summarization of documents into a few sentences. The pointer-generator network utilizes a bi-LSTM encoder and LSTM decoder with attention. At each decode step, the scalar value $p_{gen} \in (0, 1)$ is calculated with a linear layer as follows:
\begin{align}
p_{gen} = \sigma(W_c c_t + W_h h_t + W_x x_t)
\end{align}
\noindent where $c_t$ is the context vector (a weighted sum of encoder hidden states, using the attention distribution), $h_t$ is the decoder hidden state, $x_t$ is the decoder input, and $W_c$, $W_h$, and $W_x$ are learnable parameters. The scalar $p_{gen}$ is used to weight the relative contributions of the vocabulary distribution $P_{vocab}$ (i.e., the distribution that is sampled in a traditional sequence-to-sequence generation model) and the attention distribution $P_{attn}$ to the final distribution $P_{final}$, which is defined for each token $w$ in the union of the vocabulary and the input tokens:
\begin{align}
P_{final}(w) = p_{gen}P_{vocab}(w) + (1 - p_{gen})P_{attn}(w)
\end{align}
\noindent From this distribution, the decoder samples with beam search the next token to produce (as shown in Figure~\ref{fig:pointer-generator}). 

Our loss function was the average negative log likelihood of the generated sequence. We used the Adagrad optimizer \citep{duchi2011adaptive} with an initial learning rate of 0.15, gradient clipping of 2.0, and early stopping on a validation set to control overfitting. Word embeddings (128-dimensional) were randomly initialized and learned as the model was trained. LSTM hidden states (256-dimensional) were initialized with random uniform initialization up to magnitude 0.02. We trained with mini-batch size of 64 and decoded using beam search (beam size = 8) with a minimum decode length of 4 tokens and a maximum of 20 tokens. The input text was truncated at 150 tokens. We did not threshold the vocabulary, which was approximately 18k in size. During training we calculated ROUGE-1, ROUGE-2, and ROUGE-L scores as proxies for title quality. We report these numbers as well as readability and relevance scores provided by human evaluators in Section~\ref{sec:experiments}.

\subsection{Reducing Token Duplication}

An artifact of using the negative log-likelihood objective function is that we often saw the same token being generated back-to-back. Intuitively, we can understand this as the algorithm being quite confident that a particular token will show up somewhere in the true title, but because there are many ways to paraphrase a functionally equivalent title, it does not know precisely where to place that token to minimize loss. Consequently, it generates the same token in multiple places to increase its chance of being correct in one of them (e.g., it generates ``Highest Salaries NBA NBA NBA'' instead of ``Highest Salaries in the NBA'' or ``Highest Salaries of NBA Players''). While this may decrease the loss function, it also decreases readability.

The original pointer-generator network paper proposed a \emph{coverage} mechanism for reducing duplication of phrases in generated summaries \cite{see2017get}. The coverage mechanism works by summing the attention distribution over all previous decoder timesteps and penalizing those tokens with high values. We find that with titles being much shorter than article summaries, a simple heuristic serves the same purpose while simplifying the model and requiring fewer parameters to learn: we simply forbid the algorithm from generating the same token twice in one title by zeroing out its probability in the final distribution in all future decode steps. While this heuristic is not foolproof (e.g., ``La La Land Awards''), we find that over 95\% of the titles in our dataset have no repeated tokens, and many with repeated tokens can easily be paraphrased by the model to avoid the repetition while remaining relevant (e.g., ``List of Mayors of Chicago'' $\rightarrow$ ``List of Chicago Mayors''). Including this heuristic yielded an instant 4.5 point boost in ROUGE score. With more training data, a more sophisticated approach for reducing token duplication may yield further improvements.

\section{Dataset Creation}
\label{sec:datasets}

\subsection{Crowdsourced Dataset}
\label{sec:crowdsourcing}
The crowdsourced dataset consists of 10,102 web tables scraped from the tables returned as featured snippets~\cite{sullivan2017answers} to user queries on Google over a span of five months from January-May 2017.

The tables were each shown in their original context to three trained crowdworkers, who were asked to provide a descriptive title for that table and to mark whether that title occurred verbatim anywhere on the page or was composed. Based on a manual inspection of the candidate titles for 100 random tables in our dataset, we found qualitatively that the most informative and relevant title (of the three provided by crowdworkers) was composed 83\% of the time. The three candidate titles for each table were then aggregated using the following heuristic: if two or more of the title are identical, accept that title; otherwise, select the title with the most tokens (since the most common failure mode for generated titles was a lack of detail). From a sample of 50 tables, this heuristic chose what we considered the best title 67\% of the time. Finally, the titled dataset was split 80/10/10 into train/validation/test splits.

The dataset contains tables from 1384 different domain names, where two major domains were wikipedia.org (72.6\%), and espn.com (0.8\%). The dataset covers diverse topics including news (e.g., bbc.co.uk, times.com), health (e.g., webmd.com, nih.gov), entertainment (e.g., boxofficemojo.com, allmusic.com), and corporations (e.g., microsoft.com, ibm.com). The average accepted title length is 40.9 characters or 6.8 tokens. The total vocabulary size is 17,862.

\subsection{Heuristic-based Datasets}
\label{sec:heuristics}
The first heuristic-based approach we tried was proposed in \cite{chirigati2016knowledge}: mining \textit{query logs} for queries that led to pages with tables on them and using the queries as candidate titles for those tables. In addition to general quality issues with the queries as titles, we found it difficult to reliably link a given query to the precise table on a page it is most relevant to when multiple tables are found on the same page. Restricting the dataset to those tables that are the dominant tables on a page biased the dataset toward tables that are the central topic of the page, introducing an over-reliance on the page title as a source of title tokens.

Next we tried using \textit{table captions} as titles for those tables that had them, but as discussed in Section~\ref{sec:compositionality} and demonstrated in Figure~\ref{fig:compositionality}, these often contained only the last link in chain of necessary text spans for generating a good title. Hoping to identify a set of more specific captions, we tried filtering the dataset to \textit{captions with three or more entities}. This had the result of overrepresenting certain domains (e.g., scientific papers, code documentation, large online catalogs) that dominated the training set. With all of these caption-based approaches, we faced the additional issue that by removing the caption from the training input (to prevent the model from learning to simply copy over the caption word-for-word), we sometimes removed the only mention of essential information required to generate a proper title. Our last attempt was to use as titles the \textit{table captions with non-stopwords from multiple different sources} on the page and all of whose tokens could be found in at least one other input field. This, unfortunately, resulted in a dataset that was too small to use.

From these attempts, it became clear to us that given the immense variety of web tables, we would be hard-pressed to find any heuristic that could generate a training set of sufficient size without introducing significant harmful bias. We consequently commenced with collecting the crowdsourced dataset described above.

\begin{table*}[th]
  \centering
  \begin{tabular}[b]{lll|lll}
    \toprule
    \textbf{Model}  & \textbf{Relevance}  &  \textbf{Readability} & \textbf{ROUGE-1} & \textbf{ROUGE-2} & \textbf{ROUGE-L}\\
    \midrule
    Page Title      & 2.25          &  2.41  & 0.510 & 0.369 & 0.461 \\
    Section Heading & 2.29          &  2.56  & 0.476 & 0.315 & 0.411 \\
    \midrule
    Generate Only   & 1.08          &  2.25  & 0.168 & 0.064 & 0.151 \\
    Copy Only       & 1.97          &  1.26  & 0.384 & 0.221 & 0.240 \\
    Copy + Generate & \textbf{2.56} &  \textbf{2.66}  & \textbf{0.647} & \textbf{0.485} & \textbf{0.574} \\
    \midrule
    Crowdsourced    & 2.72          &  2.74  &  &  &  \\
    \bottomrule
  \end{tabular}
  \bigskip
  \caption{Comparison of relevance and readability (from human evaluation) and ROUGE metrics for 200 titles. Results are grouped by category (baselines, neural networks, crowdsourced) and shown graphically in Figure~\ref{fig:results}. The crowdsourced titles have no ROUGE scores because they are the ground truth.} 
  \label{tab:results}
\end{table*}

\section{Experimental Evaluation}
\label{sec:experiments}

We validated the quality of our approach with a human evaluation on 200 tables from the held-out test split of the dataset. We created two baselines that simply select the page title or nearest section heading, respectively, with a few obvious post-processing steps, such as removing ``[edit]'' from wikipedia section headings. We ran the pointer-generator network in \emph{Copy Only} mode ($p_{gen}$ hardcoded to 0 so no tokens are generated), \emph{Generate Only} mode ($p_{gen}$ hardcoded to 1 so all tokens are generated, as in a standard sequence-to-sequence model), and without modification (\emph{Copy + Generate}). Finally, we also evaluated the quality of the best crowdsourced title for each table as an effective upper bound. 

Evaluators assessed the readability and relevance of each title on a 3-point scale with the following interpretations: for readability, 1~=~Poor, 2~=~Medium, 3~=~Well. For relevance, 1~=~Needs Not Met, 2~=~Needs Somewhat Met, 3~=~Needs Fully Met. Experimental results are shown in Table~\ref{tab:results}. 

Prior to collecting a dataset via crowdsourcing (Sec.~\ref{sec:crowdsourcing}), we attempted to bootstrap a training dataset using four different heuristics (Sec.~\ref{sec:heuristics}). In each case, we found that the heuristic we used biased the training data enough to significantly hinder its ability to generalize. Consequently, the results reported in this paper all correspond to models trained with the crowdsourced data.
Interestingly, while \emph{Copy + Generate} had the strongest performance, \emph{Copy Only} and \emph{Generate Only} had the weakest, with each performing much better in one of the two metrics than the other, as shown in Figure~\ref{tab:results}. Since both readability and relevance are required in any real-world use case, it makes sense that historically, simpler selection-based approaches have been preferred to generation-based ones. Combining the two mechanisms, however, results in a model that performs better than both single-mechanism models in all metrics.

\textbf{Generate Only:} We observe that the \emph{Generate Only} model produced much more readable output than \emph{Copy Only}, but scored extremely low on relevance. The failure of \emph{Generate Only} seems to stem from the fact that because of the prevalence of rare and OOV tokens, the model was unable to learn sufficiently precise embeddings to convey the required information to the decoder. It did, however, appear to capture general topics. For example, when trying to produce the title ``Jyotii Sethi Filmography,'' it produced ``List of Movies with Sue Perkins'' instead---both describe a list of a particular actress's movies, but clearly the model lacked the exposure to the tokens ``Jyotii Sethi'' to either include them in its vocabulary or learn precise enough embeddings for them.

\textbf{Copy Only:} On the other hand, the \emph{Copy Only} model produced more relevant output---often producing very specific words pertaining to the table at hand---but had a much harder time composing text snippets in a readable way. Intuitively, this is because the metadata (e.g., section headings, column headers) associated with semi-structured data such as lists and tables often lack the prepositions, conjunctions, and other transition words necessary to combine multiple pieces of information. As a result, titles often consisted of snippets concatenated in unnatural ways.

\textbf{Copy + Generate:} Examining the output of the \emph{Copy + Generate} model, we found many instances where a combination of both copied tokens and generated tokens was used to great effect. For example, many tables in our dataset report movie or television appearances of an individual. A common title pattern for these is ``Ash King Filmography'' (if the table is about Bollywood singer Ash King). Instead, for that table, the \emph{Copy + Generate} model produced the title ``Filmography of Ash King''. While it may have been a mistake to generate ``Filmography'' first in the title, because it can generate filler words and had learned a reasonable language model, the model was able to recover from this misstep by adding the preposition ``of'', which actually did not occur anywhere in the input. This simple example illustrates the power of combining these two mechanisms in a single model.

It is worth noting that while the absolute difference between \emph{Copy + Generate} and the \emph{Section Heading} baseline is not large, the learned model cuts the gap to human-generated performance by more than 50\% in both readability and relevance. This corresponds to a quite noticeable difference from a reader's standpoint. Additionally, the task of generating focused natural language is one in which performing moderately is not too difficult, but performing well enough to produce genuinely usable text is very challenging.

\begin{figure}[t]
  \includegraphics[width=3in]{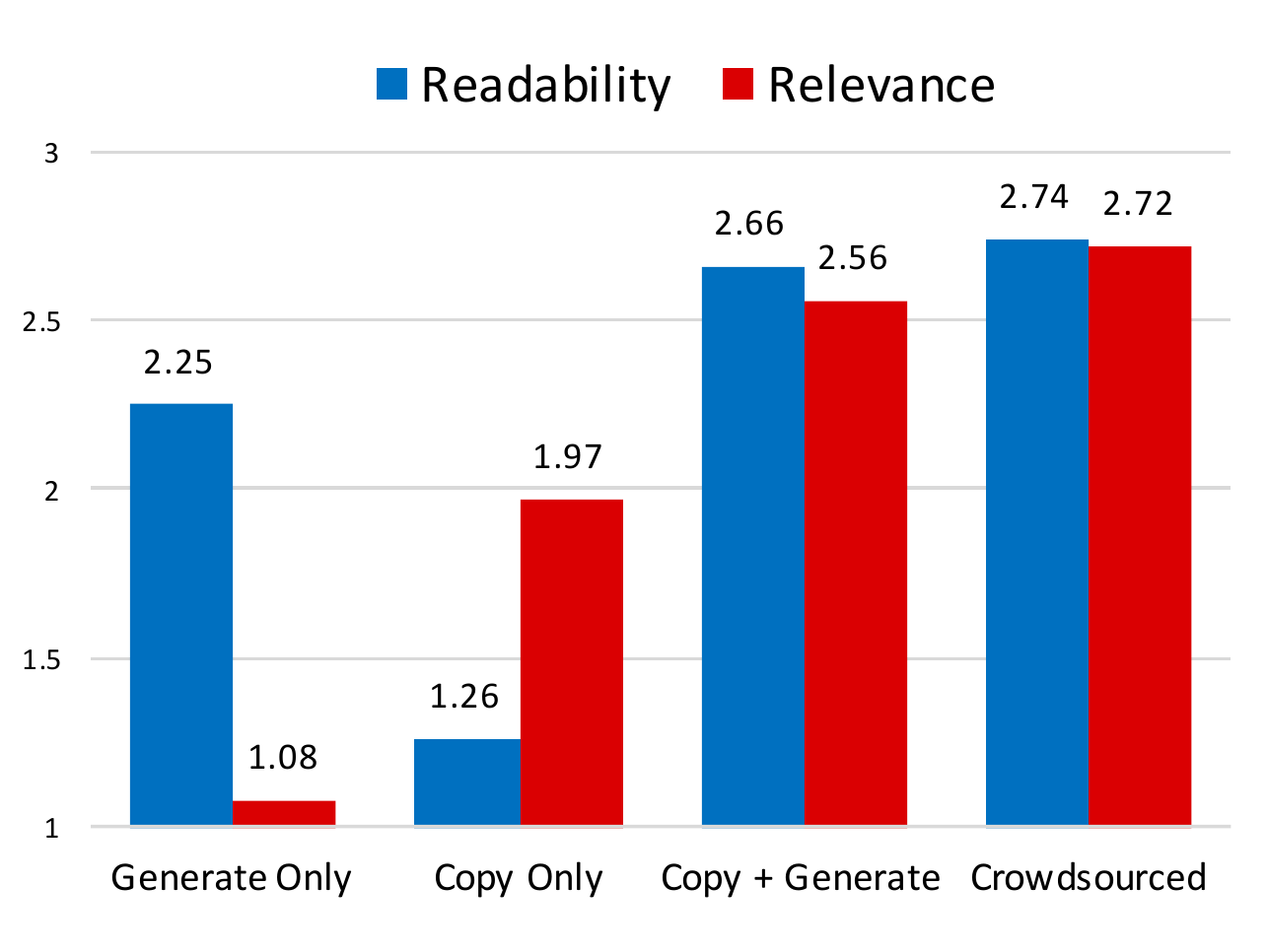}
  \caption{The poor performances of the \emph{Generate Only} and \emph{Copy Only} variants highlight the necessity of having both mechanisms in a single model to produce high quality titles.}
  \label{fig:results}
\end{figure}

\section{Related Work}
\label{sec:related}
The approach of flattening structured fields into a sequence was also recently taken by \cite{chisholm2017learning}, who used it for the related task of generating one-sentence biographies from Wikidata facts. Instead of employing a copy mechanism, they add an autoencoder objective to their sequence-to-sequence model to encourage relevance in generated text. Other attempts to summarize structured web content include generating text summaries from individual knowledge base triples \cite{vougiouklis2018neural}, generating descriptions of restaurants from structured attributes using rule-based \cite{puzikov2018e2e} or learned \cite{wiseman2018learning} templates, and generating titles for web pages with e-commerce search results given a set of attributes shared by all items on the page \cite{mathur2017generating}.

In addition to pointer-generator networks, multiple other types of networks with copy mechanisms have been proposed \citep{vinyals2015pointer,gu2016incorporating,gulcehre2016pointing, li2018point}, each with minor differences affecting when and how the network chooses to generate versus copy. 

Another way to attempt to deal with rare and OOV words is to learn over sub-word units, using character-level embeddings \cite{lee2016fully,chung2016character,luong2016achieving}, byte-pair encodings \cite{sennrich2015neural}, or WordPieces \cite{wu2016google}, for example. One reason why these methods often work well is because they are somewhat robust to variation in prefixes or suffixes of words, making it easier to recognize the similarity between words like ``slower'' and ``slowly'' compared to a word-based approach with independent embeddings for each. In the table title domain, however, we found that the most common source of OOV words was uncommon named entities, which reduces the benefit of sub-word embeddings.

For long document summarization, \cite{paulus2017deep} also reduced repetition by forbidding duplicate n-grams (in their case, trigrams). They also add decoder attention (to increase awareness of which tokens have already been generated) and intra-temporal attention on the input (similar to the coverage mechanism of \cite{see2017get}), which penalizes input tokens that have received high attention scores in the past in hopes of limiting the number of time steps that a given token can influence.

The question of how to best evaluate Natural Language Generation (NLG) systems is one of ongoing debate. While automated methods such as BLEU and ROUGE scores are fast and cheap, these have been found to show surprisingly weak correlation with human ratings \cite{novikova2017we} for many tasks and domains. (We see this in our own dataset as well, where the \emph{Page Title} baseline outperforms \emph{Section Heading} in all three ROUGE metrics, yet underperforms it in both human evaluation metrics). In general, human evaluation is still the preferred method for assessing final system quality where possible \cite{belz2006comparing}. Nevertheless, as they are a standard point of comparison, we report ROUGE scores as well.

\balance

\section{Future Work}
\label{sec:future}

Surprisingly, the pointer-generator network produces effective titles when trained with just 8k examples, a much smaller dataset than is typically used in text generation tasks. Empirically, this appears to be because the model focuses more on learning when and where to ``copy and paste'' than it does on learning accurate embeddings for all words in the vocabulary. Using the fact that words of similar parts of speech or entity types will tend to be used in the same ways in titles, in follow-up work we intend to explore the benefits of injecting extra syntactic/semantic information (such as part of speech or named entity tags) directly into the input stream. We anticipate that this will further improve performance while maintaining low training data requirements.

As we reported in this paper, we found prefix text and suffix text to have too low signal-to-noise ratios (i.e., too many irrelevant words compared to relevant ones) to justify their presence in the input to our model. However, certain key phrases in these fields seem to strongly indicate that the surrounding words will be relevant for the title---phrases such as ``the following table'' and ``as shown below''. We suspect that a model better equipped to handle long input sequences (such as an LSTM with skipping \cite{yu2017learning} or a Transformer \cite{vaswani2017attention}) may be able to better utilize fields such as these, resulting in a richer input sequence from which to generate a title. Additionally, one might take advantage of the increasingly prevalent structured metadata embedded in webpages in formats such as RDFa, Microdata and JSON-LD.

\section{Conclusion}

In this work, we presented a new framework for generating titles for web tables. We found that for the majority of web tables, the ideal title must be composed, rather than selected, from the text snippets associated with the table. We trained a sequence-to-sequence model to generate titles from input sequences consisting of the flattened key-value pairs of metadata associated with web tables. The best performing model utilized a copy mechanism to make titles relevant (often copying rare and OOV words directly from the input), and a generation mechanism to make titles readable (drawing on a learned language model for titles). In a human evaluation, the model with both copy and generation mechanisms outperformed all other models, approaching the quality of crowdsourced titles. We believe this establishes a new state of the art for table title generation.

\begin{acks}
The authors would like to thank Quoc Le and Jialu Liu for their helpful feedback over the course of this project.
\end{acks}

\newpage

\bibliographystyle{ACM-Reference-Format}
\bibliography{references} 

\end{document}